\documentclass[letterpaper, 10 pt, conference]{ieeeconf}  

\IEEEoverridecommandlockouts                              

\usepackage{amsmath,amsfonts,amscd}
\usepackage{float}
\usepackage{graphicx}
\usepackage{epstopdf}
\usepackage{color}
\usepackage{verbatim}
\usepackage{tikz,times}
\usepackage{xcolor}
\usepackage{lipsum}
\usepackage{makecell}
\usepackage{booktabs}
\usepackage{algorithm}
\usepackage{algpseudocode}
\usepackage{tabularx}
\usepackage{cellspace}
\usepackage[Export]{adjustbox}
\usepackage{caption}
\usepackage{cuted}
\usepackage{pifont}
\usepackage{bbm}
\usepackage{gensymb}

\newcommand{\cmark}{\ding{51}}%
\newcommand{\xmark}{\ding{55}}%
\newcommand{\T}{^{\mathsf{T}}} 
\newcommand{\vb}{\boldsymbol}

\makeatletter
\let\NAT@parse\undefined
\makeatother
\usepackage{hyperref}
\hypersetup{
  colorlinks = true,
  linkcolor = red,
  anchorcolor = red,
  citecolor = cyan,
  urlcolor = cyan,
  pdftitle={MM3DGS SLAM}
}
\usepackage[capitalize]{cleveref}

\title{\LARGE \bf
MM3DGS SLAM: Multi-modal 3D Gaussian Splatting for SLAM Using Vision, Depth, and Inertial Measurements
}

\author{Lisong C. Sun, Neel P. Bhatt, \IEEEmembership{Member, IEEE}, Jonathan C. Liu, Zhiwen Fan, \\Zhangyang Wang, \IEEEmembership{Senior Member, IEEE}, Todd E. Humphreys, \IEEEmembership{Senior Member, IEEE}, and Ufuk Topcu, \IEEEmembership{Fellow, IEEE}
\thanks{Lisong C. Sun, Neel P. Bhatt, Jonathan C. Liu, Zhiwen Fan, Zhangyang Wang, Todd E. Humphreys, and Ufuk Topcu are with the University of Texas at Austin, Austin, TX, USA. Email: {\tt\small \{codey.sun, npbhatt, jonathanliu88, zhiwenfan, atlaswang, todd.humphreys, utopcu\}@utexas.edu}}%
}

\begin{document}

\maketitle
\thispagestyle{empty}
\pagestyle{empty}

\begin{strip}
\begin{minipage}{\textwidth}\centering
\vspace{-57pt}
    \includegraphics[width=0.85\linewidth]{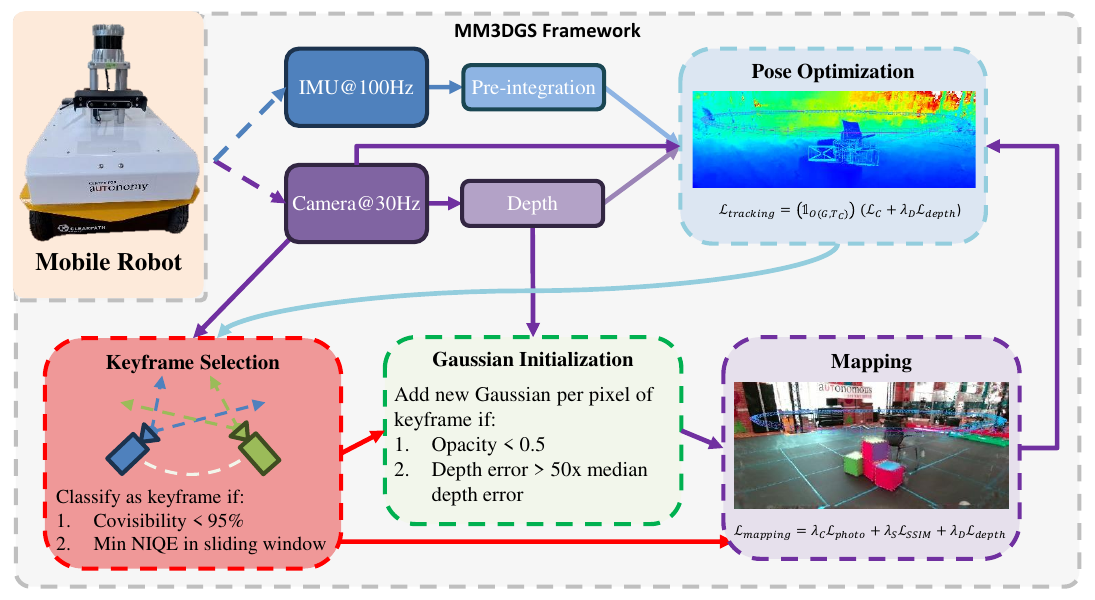}
    \captionof{figure}{\small Overview of the MM3DGS framework. We receive camera images and inertial measurements from a mobile robot. We utilize depth measurements and IMU pre-integration for pose optimization using a combined tracking loss. We apply a keyframe selection approach based on image covisibility and the NIQE metric across a sliding window and initialize new 3D Gaussians for keyframes with low opacity and high depth error \cite{mittal2013niqe}. Finally, we optimize parameters of the 3D Gaussians according the mapping loss for the selected keyframes.}
    \label{fig:overview}
\end{minipage}
\end{strip}

\begin{abstract}

Simultaneous localization and mapping is essential for position tracking and scene understanding. 3D Gaussian-based map representations enable photorealistic reconstruction and real-time rendering of scenes using multiple posed cameras. We show for the first time that using 3D Gaussians for map representation with unposed camera images and inertial measurements can enable accurate SLAM. Our method, MM3DGS, addresses the limitations of prior neural radiance field-based representations by enabling faster rendering, scale awareness, and improved trajectory tracking. Our framework enables keyframe-based mapping and tracking utilizing loss functions that incorporate relative pose transformations from pre-integrated inertial measurements, depth estimates, and measures of photometric rendering quality. We also release a multi-modal dataset, UT-MM, collected from a mobile robot equipped with a camera and an inertial measurement unit. Experimental evaluation on several scenes from the dataset shows that MM3DGS achieves 3× improvement in tracking and 5\% improvement in photometric rendering quality compared to the current 3DGS SLAM state-of-the-art, while allowing real-time rendering of a high-resolution dense 3D map.

\end{abstract}

\section{INTRODUCTION}
{\let\thefootnote\relax\footnote{Webpage: \url{https://vita-group.github.io/MM3DGS-SLAM}}}Simultaneous localization and mapping (SLAM), the task of generating a map of the environment along with estimating the pose of a sensor, is an essential enabler in applications such as aerial mapping, augmented reality, and autonomous mobile robotics \cite{contreras2023stereo, polvi2016slidar}. 3D scene reconstruction and sensor localization are essential capabilities for an autonomous system to perform downstream tasks such as decision-making and navigation \cite{bavle2020vps}. As a result, SLAM plays a pivotal role in advancing the capabilities of autonomous systems.

The representation used for mapping and the type of sensor input utilized is a fundamental choice that has a direct impact on SLAM performance. Although SLAM approaches using sparse point clouds to represent the operating environment yield state-of-the-art tracking accuracy \cite{mur2015orb,leutenegger2015keyframe}, the generated maps are disjoint due to sparsity and visually inferior to newer 3D reconstruction methods. While visual quality is irrelevant for the sole purpose of navigation, the creation of photorealistic maps is valuable for human consumption, semantic segmentation, and post-processing. Neural radiance field (NeRF)-based SLAM approaches yield output maps that are spatially and photorealistically dense \cite{sucar2021imap,zhu2022nice}, but are computationally expensive at inference time and hence are not capable of accurately tracking in real-time. To address this shortcoming, recent works utilize the real-time rendering capability of a 3D Gaussian-based map representation and perform 3D Gaussian splatting (3DGS) to generate 2D renderings \cite{matsuki2024gsslam, keetha2023splatam}.

A precursor to 3DGS is depth initialization of the Gaussians. Most approaches use depth inputs from relatively expensive sensors such as LiDARs, which may not be readily available on a device such as a consumer phone \cite{hong2024liv}. Other approaches use an RGB-D camera providing depth information through stereoscopic depth estimation \cite{keetha2023splatam}. However, relying solely on RGB-D cameras may lead to erroneous depth, especially as the distance from the camera increases, leading to degraded tracking accuracy.

To address these shortcomings, we present the first real-time visual-inertial SLAM framework using 3D Gaussians for efficient and explicit map representation with inputs from a single monocular camera or RGB-D camera along with inertial measurements that may be readily obtained from most modern smartphones. Our approach is capable of real-time photorealistic 3D reconstructions of the environment, yielding accurate camera pose tracking for SLAM. We release a multi-modal dataset consisting of several scenes and the required sensory inputs which we use to evaluate our approach. Our approach outperforms SplaTAM, the state-of-the-art RGB-D-based 3DGS SLAM approach, by \textbf{3x} in trajectory tracking and achieves a \textbf{5}\% increase in photorealistic rendering quality. 

In summary, our key contributions are as follows:
\begin{itemize}
    \item We integrate inertial measurements and depth estimates from an unposed monocular RGB or RGB-D camera into our real-time MM3DGS SLAM framework using 3D Gaussians for scene representation. Our framework enables scale awareness as well as lateral, longitudinal, and vertical trajectory alignment. Our framework can utilize inputs from inexpensive sensors available on most consumer smartphones.
    \item We release a multi-modal dataset collected using a mobile robot consisting of several indoor scenarios with RGB and RGB-D images, LiDAR depth, 6-DOF inertial measurement unit (IMU) measurements, as well as ground truth trajectories for error analysis. 
    \item We achieve superior quantitative and qualitative trajectory tracking (\textbf{3x} improvement) and photometric rendering (\textbf{5\%} improvement) results compared to the current state-of-the-art 3DGS SLAM baseline.
\end{itemize}

\section{RELATED WORKS}

\subsection{SLAM Map Representations}
Sparse visual SLAM algorithms, such as ORB-SLAM~\cite{mur2015orb} and OKVIS~\cite{leutenegger2015keyframe}, are designed to estimate precise camera poses while producing only sparse point clouds for map representation. Conversely, dense visual SLAM methodologies~\cite{newcombe2011dtam,schops2019bad} aim to construct a dense representation of the scene. Map representations in dense SLAM are broadly classified into two categories: view-centric and world-centric. View-centric representations encode 3D information using keyframes accompanied by depth maps~\cite{newcombe2011dtam}. On the other hand, world-centric approaches anchor the 3D geometry of the entire scene within a consistent global coordinate system, typically represented through surfels~\cite{schops2019bad} or by utilizing occupancy values within voxel grids~\cite{dai2017bundlefusion}.

\subsection{Efficient 3D Representation} \label{sec:3dgs}
Recent advancements in neural rendering, exemplified by NeRFs \cite{mildenhall2021nerf} and subsequent developments \cite{fridovich2022plenoxels,muller2022instant}, have demonstrated significant progress in novel view synthesis using foundational 3D representations such as MLPs, voxel grids, or hash tables. Although these NeRF-inspired models exhibit impressive results, they often necessitate extensive training periods for individual scenes.
The introduction of 3DGS, which this paper employs, offers a solution to the issue of training efficiency. This volumetric rendering approach depicts a 3D scene as a collection of explicit Gaussian distributions \cite{kerbl20233d}. Initially, reconstructing a scene with 3DGS required applying structure-from-motion techniques like COLMAP \cite{schoenberger2016sfm, schoenberger2016mvs} to determine camera poses before optimization, but recent studies have aimed to bypass this prerequisite by leveraging depth measurements or monocular depth estimators \cite{keetha2023splatam, fu2023colmapfree}. Concurrently, several research initiatives have explored the utilization of 3DGS within SLAM frameworks, with works in \cite{yan2024gsslam, keetha2023splatam, matsuki2024gsslam, yugay2023gaussianslam} employing RGB-D sequences. Specifically, Gaussian Splatting SLAM \cite{matsuki2024gsslam} considers both RGB and RGB-D settings, but none of these frameworks fuse inertial measurements. At the time of writing, SplaTAM, an RGB-D-based 3DGS method, is the only method with publicly available source code and is thus considered as the baseline \cite{keetha2023splatam}.

\subsection{Multi-modal SLAM Frameworks}
Visual sensing, while effective, has its limitations, such as susceptibility to motion blur and exposure changes. To mitigate the vulnerabilities of individual sensors, a standard approach is to employ multi-modal sensing through sensor fusion. Prior research has delved into enhancing system robustness by integrating semantic mapping and developing LiDAR-camera SLAM systems augmented with laser range finders~\cite{hong2024liv,jeong2018towards,jiang2016static}. In contrast, our focus is on the fusion of inertial measurements, which are favored for their high data acquisition rates and proficiency in tracking rapid movements within brief timeframes.

\section{METHOD}

The MM3DGS SLAM framework consists of four main stages: pose optimization (tracking), keyframe selection, Gaussian initialization, and mapping. An overview of the framework is illustrated in \cref{fig:overview}.

\subsection{3D Gaussian Splatting}

The underlying scene map is represented using a set of 3D Gaussians $\mathcal{G}$, where the $i$th Gaussian is defined by position $\vb{\mu}_i$, shape $\Sigma_i$, opacity $o_i$, and color $\vb{c}_i$. Recall that given a mean, $\vb{\mu}$, and covariance matrix, $\Sigma$, a Gaussian distribution is defined as

\begin{equation} \label{eq:gaussianpdf}
    G(\vb{x}) = \exp(-\frac{1}{2}(\vb{x}-\vb{\mu})\Sigma^{-1}(\vb{x}-\vb{\mu})\T)
\end{equation}

By applying eigenvalue decomposition to $\Sigma$, the covariance can be decomposed into the form, $\Sigma=RSS\T R\T$, where $R$ represents an orthonormal rotation matrix and $S$ represents a diagonal scaling matrix. In this way, the shape of 3D Gaussians can be optimized while keeping $\Sigma$ symmetrical and positive-semidefinite.

A set of 3D Gaussians can be rasterized into an image via ``splatting," i.e., projecting the Gaussians onto a 2D image plane. The 2D view-space covariance matrix, $\Sigma'$, can be computed as $\Sigma' = JW\Sigma(JW)\T$, where $J$ is the Jacobian of the affine approximation of the projection matrix, $W$ is the world to view frame transformation matrix, and $\Sigma$ is the 3D covariance matrix of the respective Gaussian.

Specifically, given a set of Gaussian features, $\mathcal{G}$, and a camera pose, $T_c$, the color, $C$, of a pixel is calculated by blending $N$ Gaussians that overlap the pixel, ordered by non-increasing depth given by

\begin{equation} \label{eq:rasterize}
    C(\mathcal{G}, T_c) = \sum_{i=1}^{N}\vb{c}_i\alpha_i\prod_{j=1}^{i-1}(1-\alpha_j)
\end{equation}

\noindent where $\vb{c}_i$ is the color of the $i$th Gaussian and $\alpha_i$ is sampled from the $i$th splatted 2D Gaussian distribution at the pixel location. Similarly, the opacity of a pixel can be calculated as

\begin{equation} \label{eq:opacity}
    O(\mathcal{G}, T_c) = \sum_{i=1}^{N}\alpha_i\prod_{j=1}^{i-1}(1-\alpha_j)
\end{equation}

Since this rendering process is differentiable, $\mathcal{G}$ can be optimized using the $L_1$ loss between the rendered image and the ground truth undistorted image, $I$:

\begin{equation} \label{eq:photoloss}
    \mathcal{L_\mathrm{photo}} = L_1(I, C(\mathcal{G}, T_c))
\end{equation}


\subsection{Tracking} \label{sec:tracking}
The tracking process consists of camera pose optimization given a fixed 3D Gaussian map. To enable gradient backpropagation to the camera pose, the inverse camera transformation is applied to the Gaussian map while keeping the camera fixed. This achieves identical rendering without implementing camera pose gradients in the 3DGS rasterizer. Subsequently, the 3D Gaussian map is frozen, and the camera pose is optimized according to the following loss function:

\begin{equation}\label{eq:tracking}
    \mathcal{L_\mathrm{tracking}} = (\mathbbm{1}_{O(\mathcal{G}, T_c)})(\mathcal{L_\mathrm{photo}} + \lambda_\mathrm{D}\mathcal{L_\mathrm{depth}})
\end{equation}

\noindent where $\mathbbm{1}_{O(\mathcal{G}, T_c)}$ is an indicator function defined as

$$
    \mathbbm{1}_{O(\mathcal{G}, T_c)} = \begin{cases}
1\hspace{0.5cm} \text{if } O(\mathcal{G}, T_c) > 0.99\\
0\hspace{0.5cm} \text{otherwise}
\end{cases}
$$

\noindent and $\mathcal{L_\mathrm{depth}}$ is the depth loss described in detail in Sec. \ref{sec:depth}. Since the map is not guaranteed to cover the entire extent of the current frame, pixels with an opacity $< 0.99$ are masked.

To aid in convergence, a dynamics model can be applied prior to optimization to provide an initial guess for the camera pose. In most cases, a constant velocity model is used. However, in the presence of an inertial sensor, tracking accuracy can be improved by utilizing inertial measurements as is later described in Sec. \ref{sec:imu}. Note that tracking is skipped for the first frame as there is no existing map yet and an identify transformation matrix is assumed as an initial guess.

\subsection{Depth Supervision} \label{sec:depth}

When tracking camera poses using a 3D Gaussian map, it is essential for the map to encode accurate geometric information absent which the tracking may diverge. This is particularly true for an underoptimized map, in which Gaussians are not trained long enough to converge to the correct position. The use of depth priors solves this problem by providing reasonable initial estimates for Gaussian positions, minimizing both inconsistent geometry and training time. Further, depth priors can supervise the map training loss to prevent geometric artifacts from overfitting on limited views.

To render depth in a differentiable manner, a second rasterization pass is performed with color $\vb{c}_i$ of each Gaussian replaced with its projected depth on the image plane.

In the RGB-D case, these depth priors can be gathered directly through a depth sensor or using stereoscopic depth. However, in the absence of a such sensors, monocular dense depth estimation networks, such as DPT \cite{ranftl2021dpt}, can be used. Since dense depth estimators output a relative inverse depth, the estimated and rendered depths cannot be directly compared. Following \cite{zhu2023FSGS}, the depth loss is instead computed using the linear correlation (Pearson correlation coefficient) between the estimated and rendered depth maps $D_e$ and $D_r$:

\begin{equation}
    \mathcal{L_\mathrm{depth}} = \frac{\mathrm{Cov}(D_e, D_r)}{\sqrt{\mathrm{Var}(D_e)\mathrm{Var}(D_r)}}
\end{equation}

\noindent This depth correlation term is appended to the loss functions used in Eqs. (\ref{eq:tracking}) and (\ref{eq:mapping}).

For initializing Gaussian positions, one must first resolve the scale ambiguity of the depth estimate. This can be done by solving for a scaling $\sigma$ and shift $\theta$ that fits the depth estimate to the current map. This can be modeled as a linear least squares problem of the form

\begin{equation}\label{depthlls}
    \begin{bmatrix}
        \vb{d_e} && \vb{1}
    \end{bmatrix}
    \begin{bmatrix}
        \sigma \\ \theta
    \end{bmatrix}
    = \vb{d_r}
\end{equation}

\noindent where $\vb{d_e}$ and $\vb{d_r}$ are the flattened vectors of $D_e$ and $D_r$. Once the estimated depths are properly fitted to the existing map, new Gaussians can be initialized for unseen areas, initializing underoptimized maps with geometric information.

\subsection{Inertial Fusion} \label{sec:imu}

Prior to optimizing the camera pose at the current frame, an initial pose estimate is required to guide optimization. A good initial estimate can lead to faster optimization times. In addition, in underoptimized areas where the convergence basins may be small, good initial estimates are essential to prevent tracking divergence.

In a monocular setting, pose estimates can be extrapolated by assuming constant velocity between consecutive frames. However, this model breaks down during presence of vigorous camera motion and low image frame rate. 

Inertial measurements obtained via an IMU can be integrated to accurately propagate the camera pose between frames and yield meaningful initial estimates. Most 6-degree-of-freedom (6-DOF) IMUs provide linear acceleration measurements through accelerometers, $\mathbf{a} = [\Ddot{x},\Ddot{y},\Ddot{z}]$, as well as angular velocities, $\mathbf{\Dot{\Theta}} = [\Dot{\alpha},\Dot{\beta},\Dot{\gamma}]$ , in 3D space. Given IMU measurements are readily available in almost all consumer phones and are relatively inexpensive compared to cameras and LiDARs, this valuable information can be practically availed in any setting.

\begin{figure}[t]
  \centering
    \includegraphics[width=.2375\textwidth]{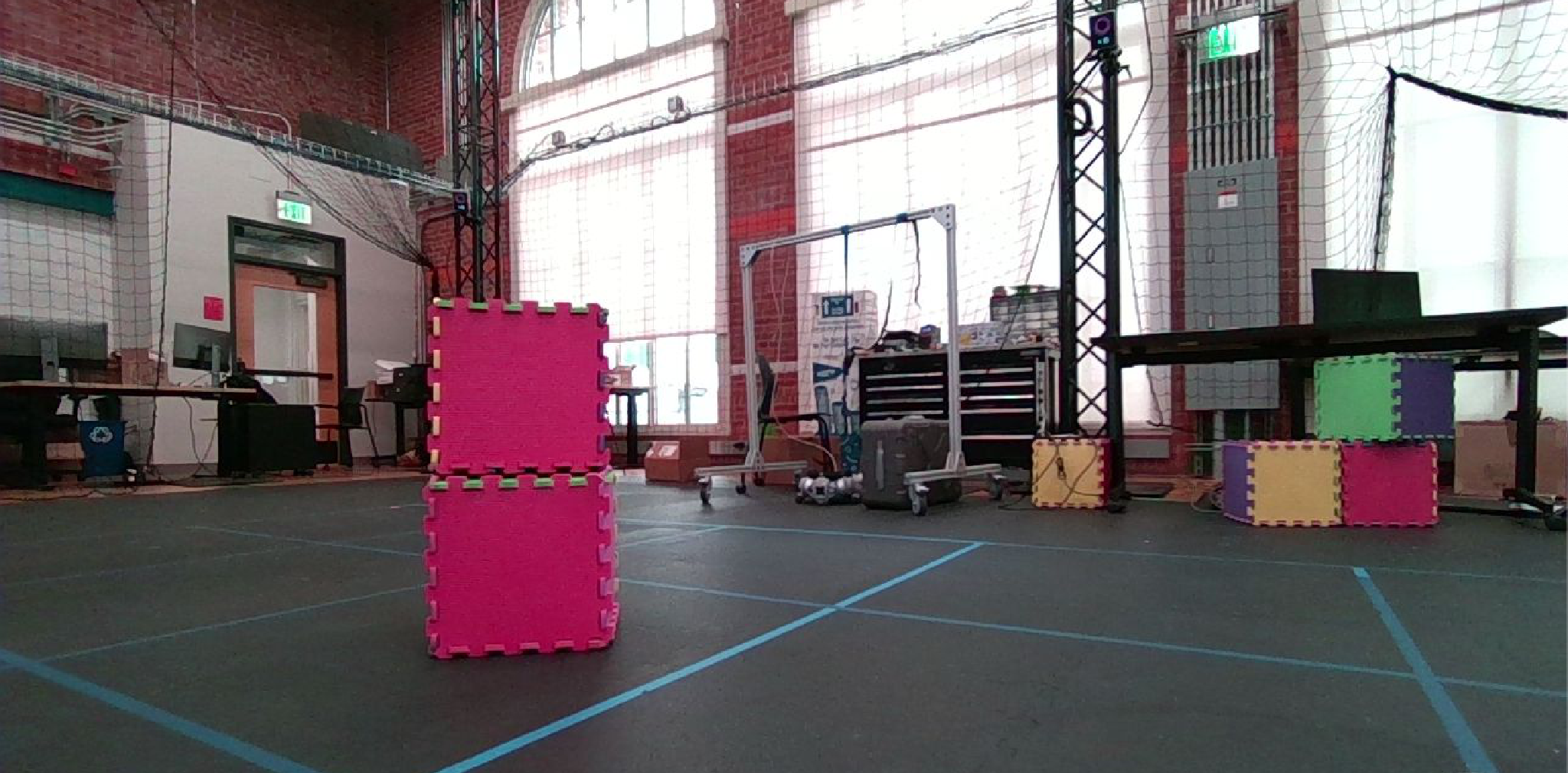}%
    \includegraphics[width=.2375\textwidth]{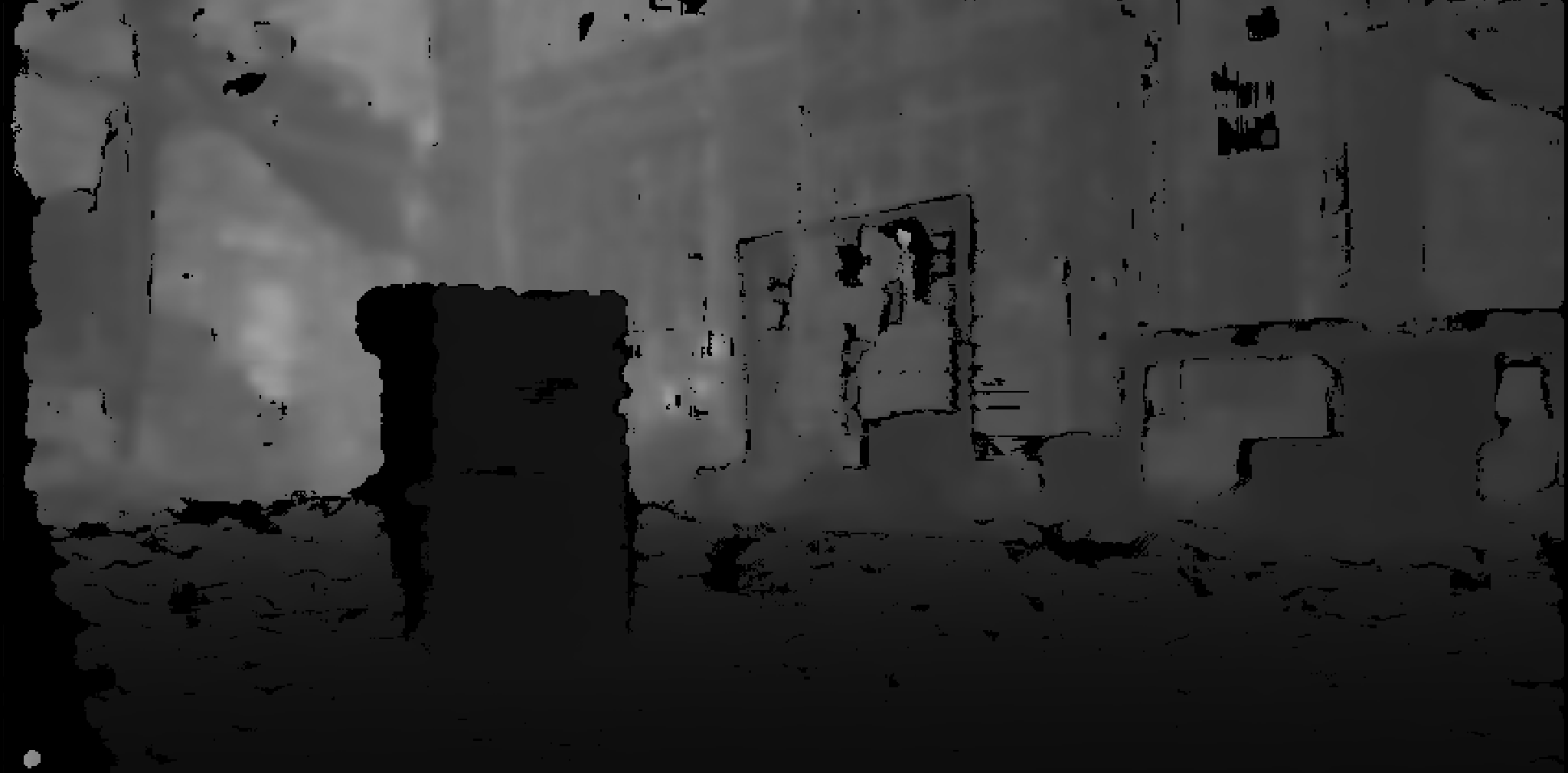}
    \includegraphics[width=.2375\textwidth]{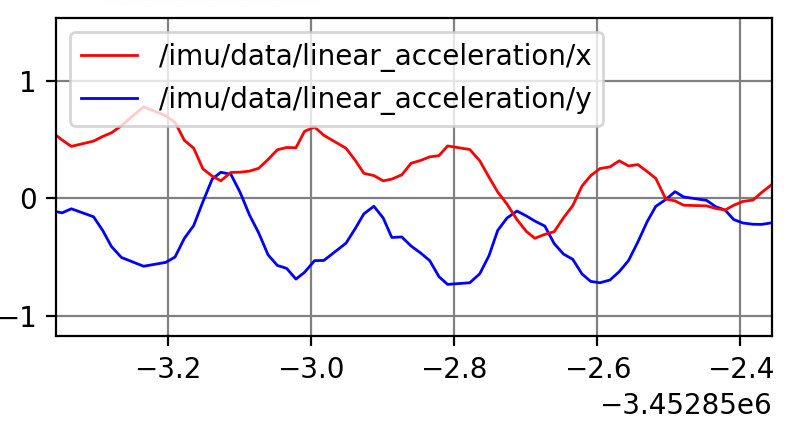}%
    \includegraphics[width=.2375\textwidth]{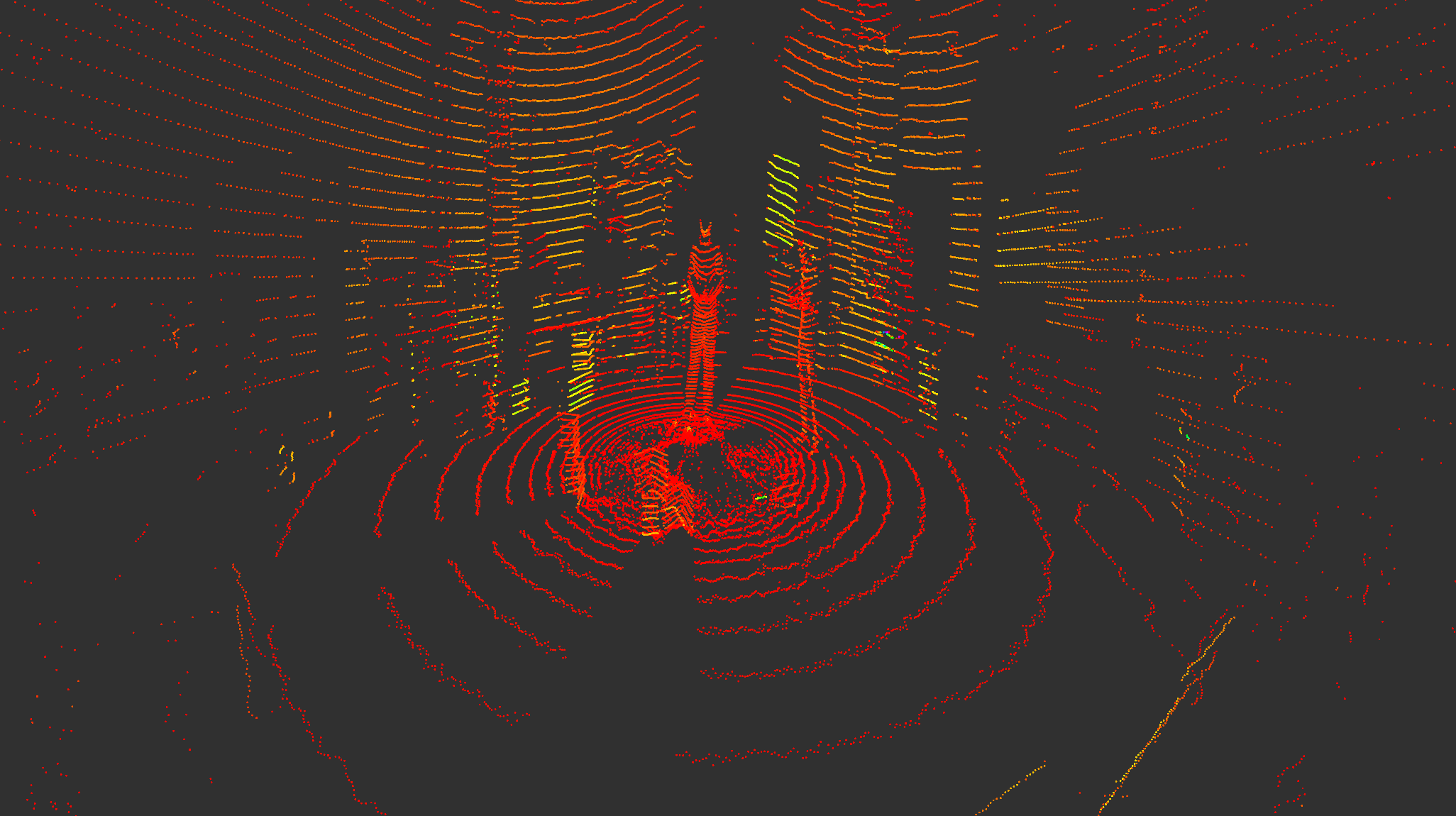}

  \caption{Our dataset provides RGB images (top left), depth images (top right), IMU measurements (bottom left), and LIDAR point clouds (bottom right). The above examples were taken from the Ego-drive scene.}
  \label{fig:UT-MM}
\end{figure}

The change in position at time, $t$, expressed in the previous coordinate frame, $^{t-1}\Delta\mathbf{p_t}$, can be computed using \cref{eq:delta_p},

\begin{equation} \label{eq:delta_p}
    ^{t-1}\Delta\mathbf{p_t} = \vb{v}_{t-1} \times t + \frac{1}{2}\mathbf{a}t^2
\end{equation}

\noindent where velocity is computed as $\vb{v}_{t-1} = \vb{v}_{t-2} + \vb{a}_{t-1} \times t$.

Similarly, the change in angular position at time, $t$, expressed in the previous coordinate frame, $^{t-1}\Delta\mathbf{p_t}$, can be computed using \cref{eq:delta_theta},

\begin{equation} \label{eq:delta_theta}
    ^{t-1}\Delta\mathbf{\Theta_t} = \mathbf{\Theta}_{t-1} \times t
\end{equation}

Using, \cref{eq:delta_p} and \cref{eq:delta_theta} the relative transformation between two consecutive coordinate frames, $^{t-1}_{t}T_{I}$, can be constructed as in \cref{eq:rel_transform},

\begin{equation} \label{eq:rel_transform}
    ^{t-1}_{t}T_{I} = [^{t-1}_{t}R \, | \, ^{t-1}\Delta\mathbf{p_t}]
\end{equation}

\noindent where the relative rotation matrix, $^{t-1}_{t}R$, is constructed using $^{t-1}\Delta\mathbf{\Theta_t}$ from \cref{eq:delta_theta}. Using the static transform between the IMU and camera frame, the relative transform between consecutive camera frames can be computed as per \cref{eq:static_transform},

\begin{equation} \label{eq:static_transform}
    ^{t-1}_{t}T_c = ^{C}_{I}T \medspace ^{t-1}_{t}T_{I}
\end{equation}

To obtain the transform between the two arbitrary frames within a sliding window, the relative transform in \cref{eq:static_transform}, can be chained from the current frame to the destination frame of interest.

Note that this open-loop method does not estimate internal IMU biases. This is because errors in $^{t-1}_{t}T_c$ are small within short time deltas, and these small errors are optimized away by the visual camera pose optimization in \cref{sec:tracking}. However, this method becomes less robust as both video frame rate and IMU quality decreases. We leave closed-loop inertial fusion with bias estimation as future work.

\subsection{Gaussian Initialization}

To cover unseen areas, new Gaussians are added each keyframe at pixels where the opacity is below $0.5$ and the depth error exceeds $50$ times the median depth error. These new Gaussians are added per-pixel, with RGB initialized at the pixel color, position initialized at the depth measurement/estimate (elaborated in Sec. \ref{sec:depth}), opacity at 0.5, and scaling initialized isotropically to cover the extent of a single pixel at the initialized depth.

\begin{figure}[t]
    \centering
    \includegraphics[width=0.43\linewidth]{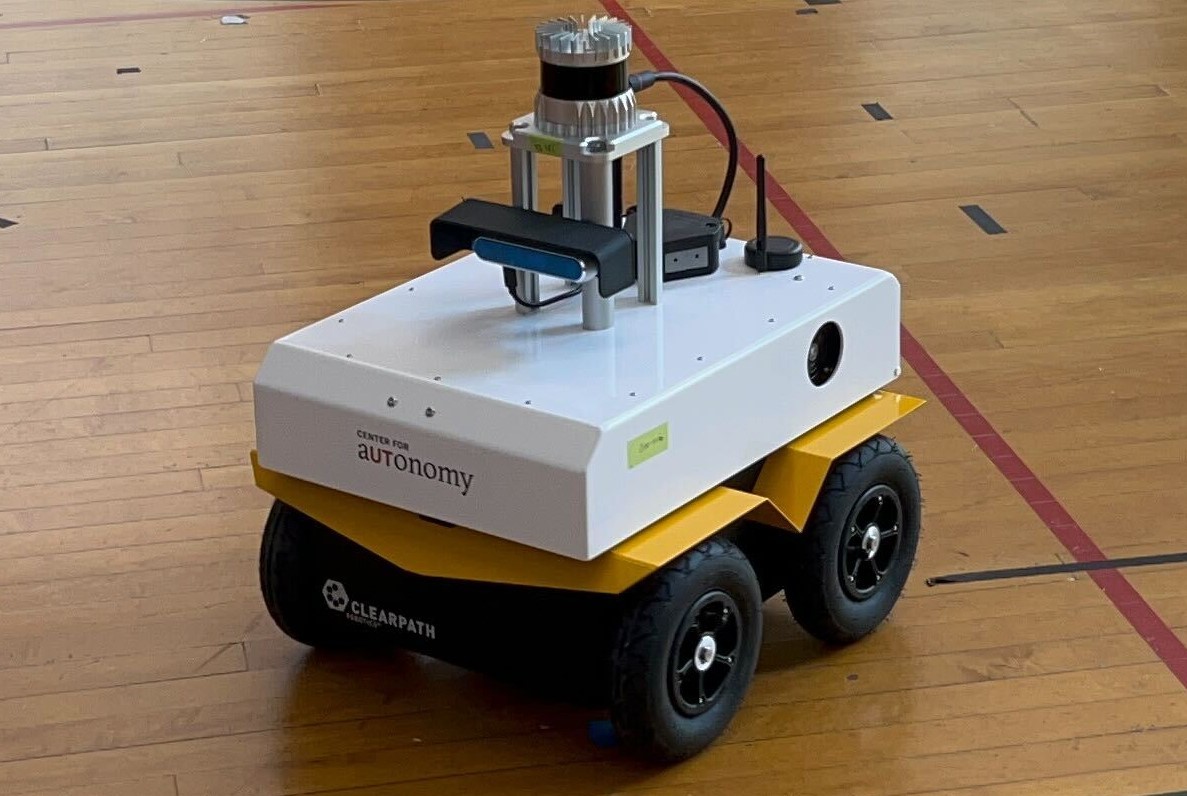}%
    \includegraphics[width=0.57\linewidth]{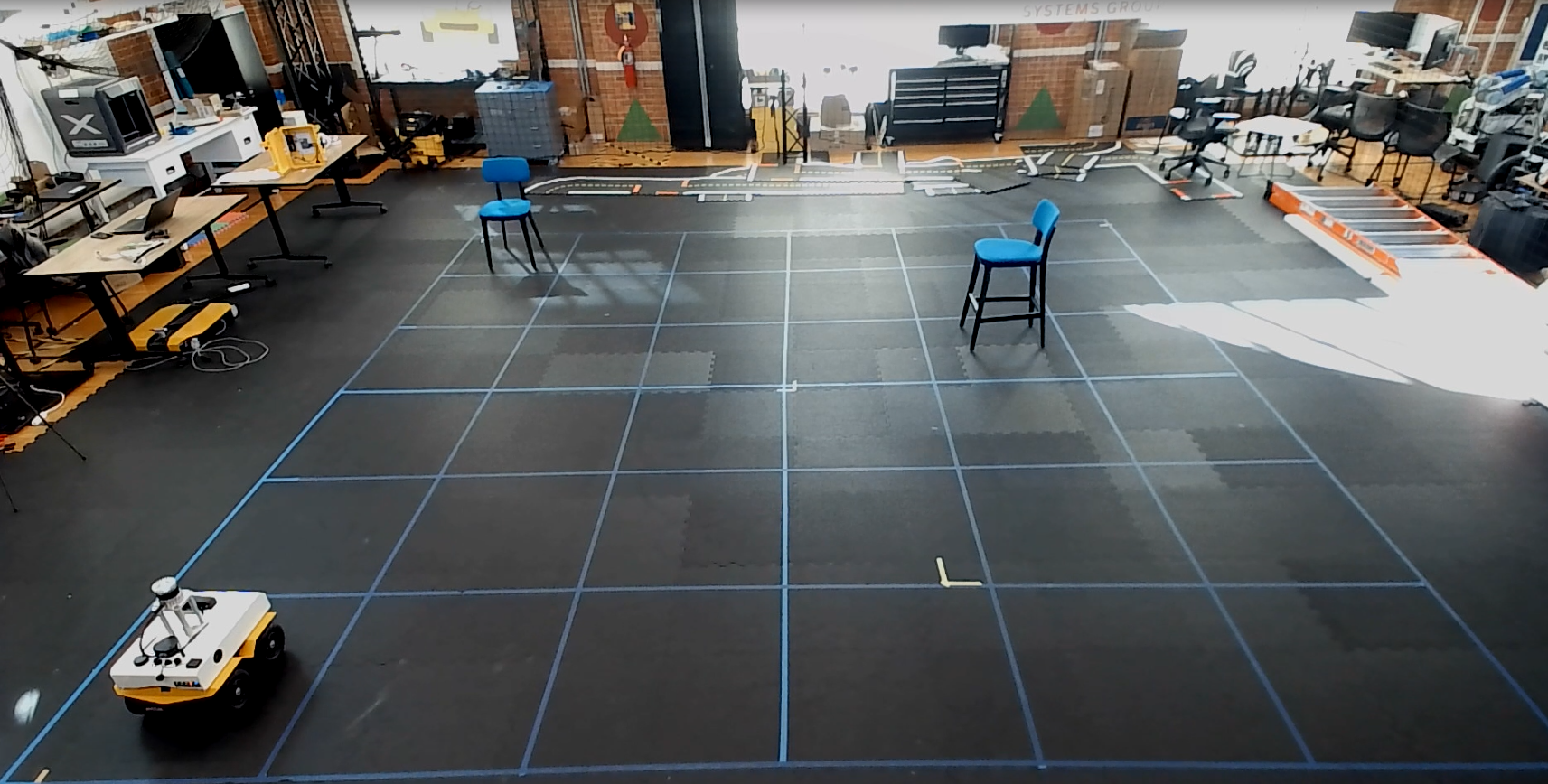}
    \caption{A depiction of the mobile robot platform (left) equipped with a RGB-D camera, IMU, and a LiDAR and the test environment (right) featuring a 16 camera Vicon-based ground truth system.}
    \label{fig:jackal}
\end{figure}

\subsection{Keyframe Selection} \label{sec:keyframe}

\begin{figure*}[t]
    \setlength\tabcolsep{0.5pt}
    \adjustboxset{width=\linewidth,valign=c}
    \centering
    \begin{tabular}{ccccccccc}
    \rotatebox[origin=B]{90}{GT}  & 
    \includegraphics[width=0.12\linewidth]{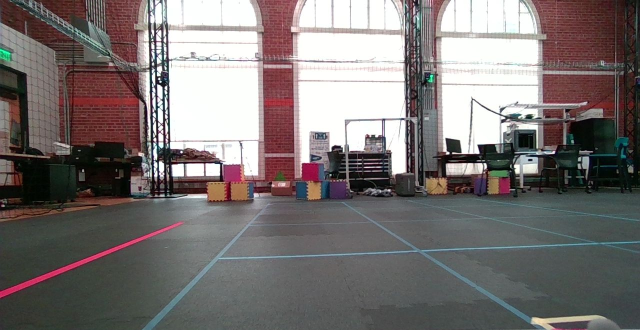} & \includegraphics[width=0.12\linewidth]{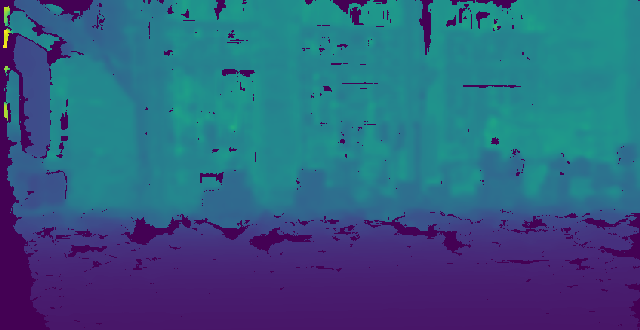} & \includegraphics[width=0.12\linewidth]{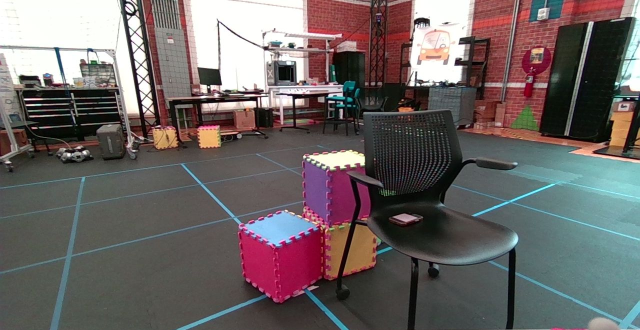} & \includegraphics[width=0.12\linewidth]{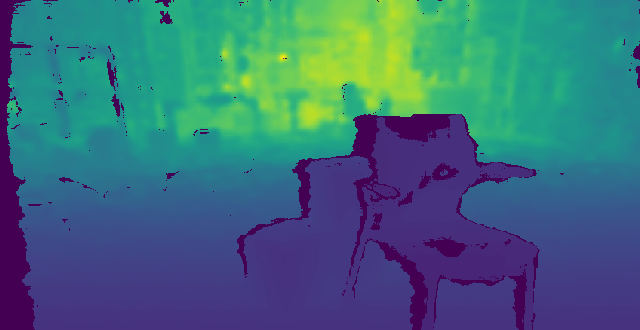} & \includegraphics[width=0.12\linewidth]{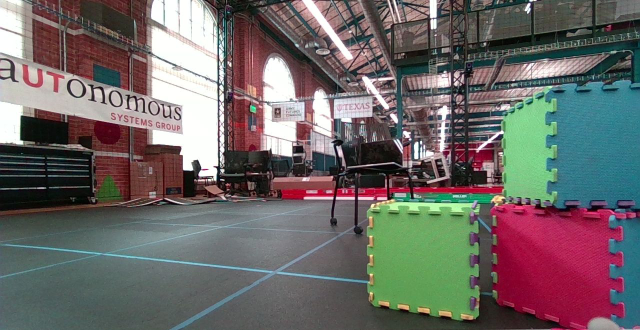} & \includegraphics[width=0.12\linewidth]{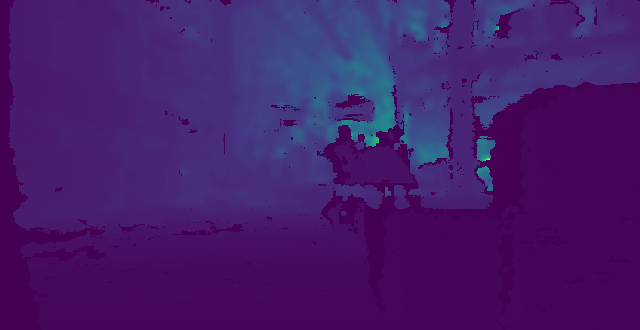} & \includegraphics[width=0.12\linewidth]{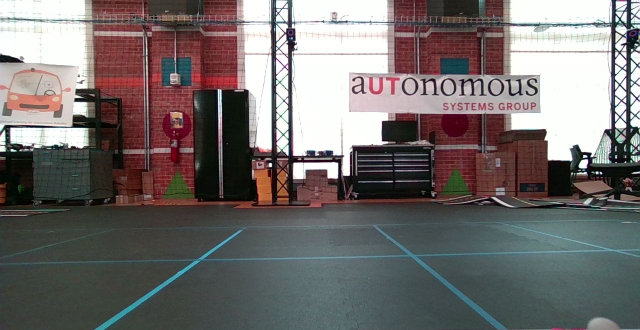} & \includegraphics[width=0.12\linewidth]{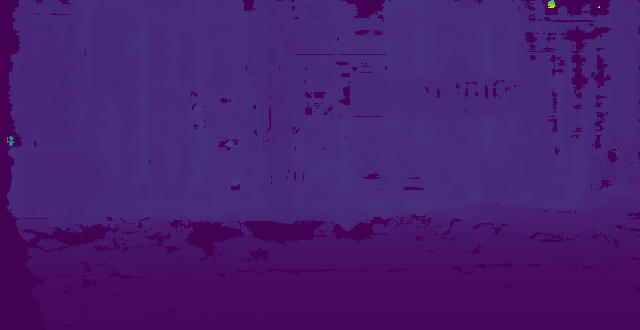} \\
    \rotatebox[origin=B]{90}{\footnotesize SplaTAM}  & 
    \includegraphics[width=0.12\linewidth]{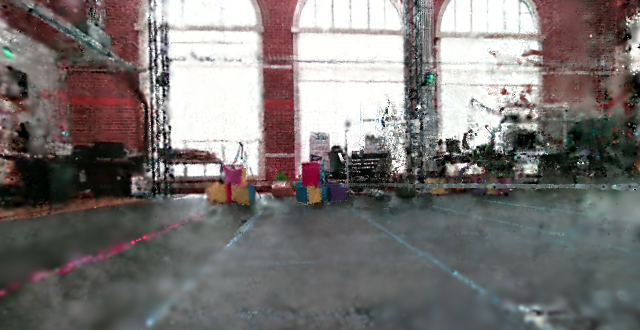} & \includegraphics[width=0.12\linewidth]{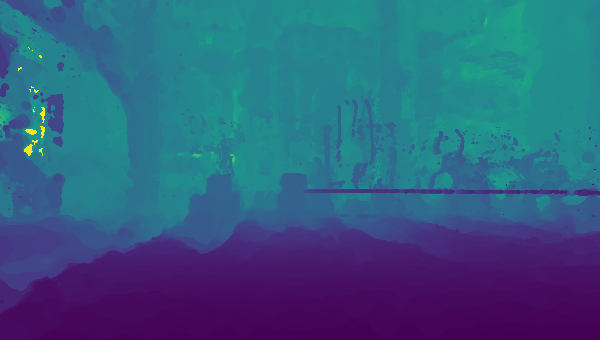} & \includegraphics[width=0.12\linewidth]{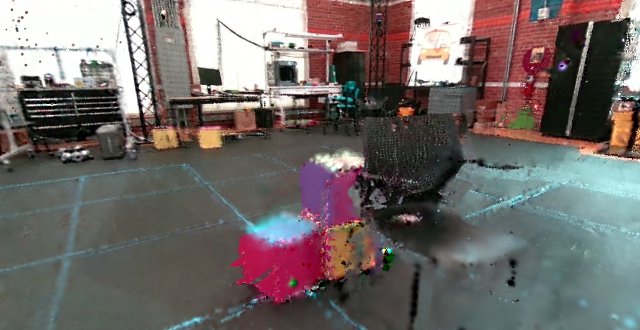} & \includegraphics[width=0.12\linewidth]{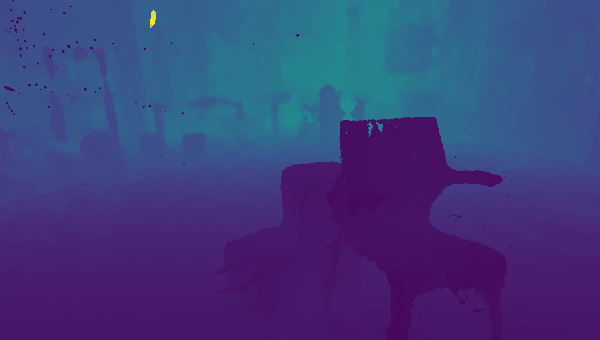} & \includegraphics[width=0.12\linewidth]{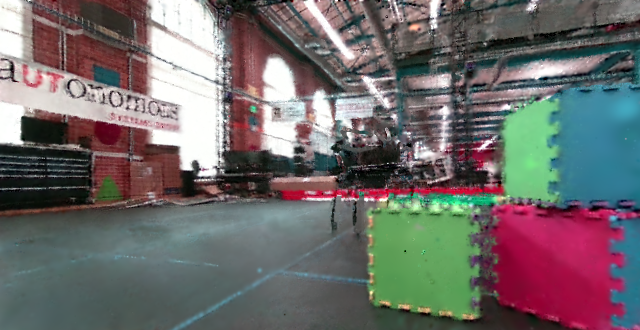} & \includegraphics[width=0.12\linewidth]{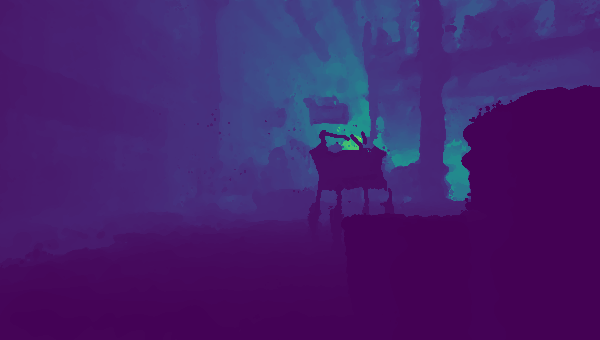} & \includegraphics[width=0.12\linewidth]{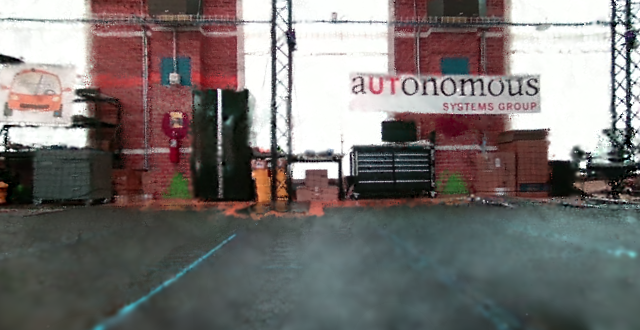} & \includegraphics[width=0.12\linewidth]{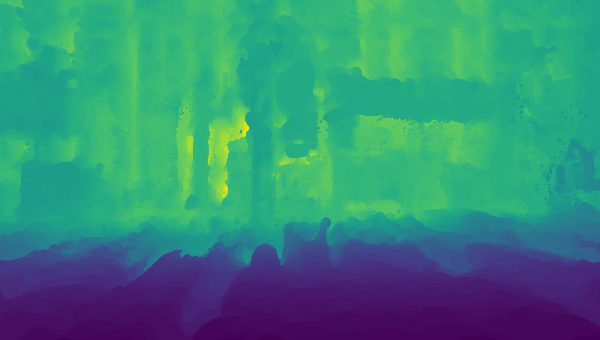} \\
    \rotatebox[origin=B]{90}{\textbf{Ours}}  & 
    \includegraphics[width=0.12\linewidth]{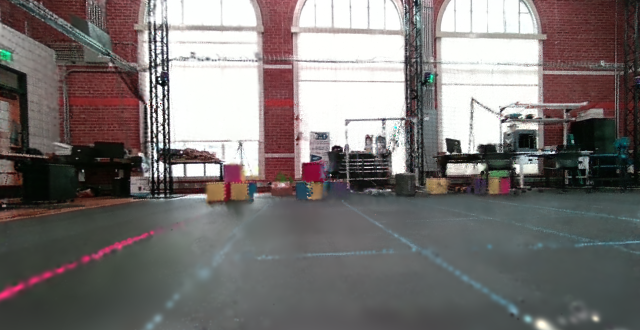} & \includegraphics[width=0.12\linewidth]{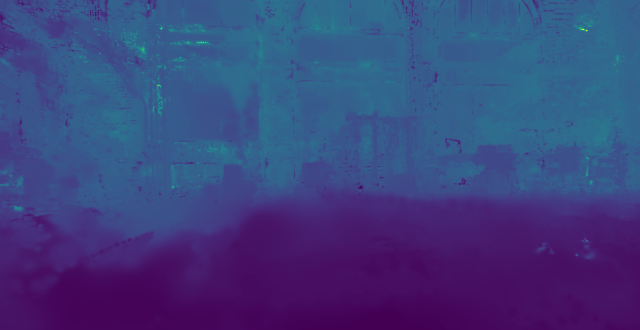} & \includegraphics[width=0.12\linewidth]{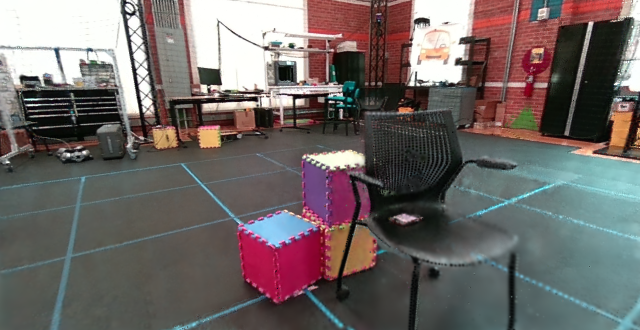} & \includegraphics[width=0.12\linewidth]{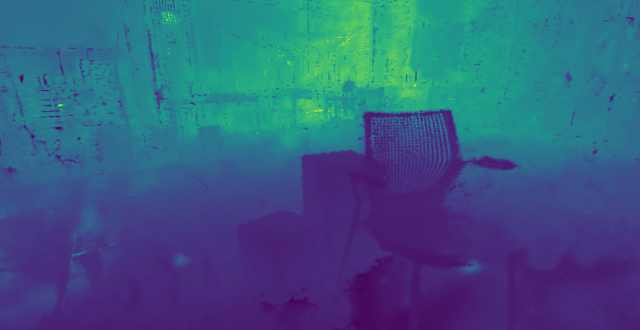} & \includegraphics[width=0.12\linewidth]{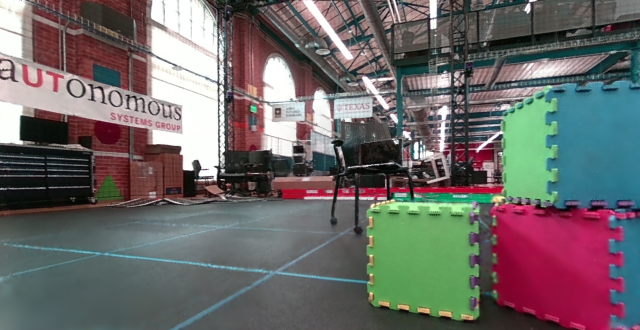} & \includegraphics[width=0.12\linewidth]{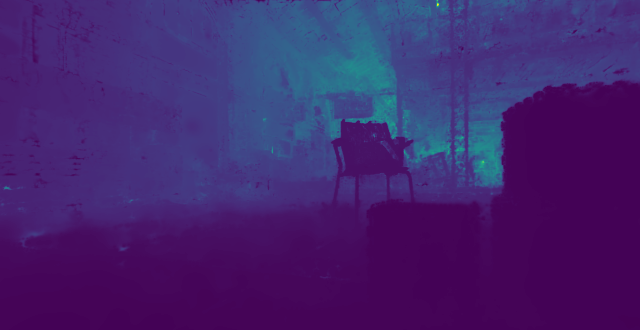} & \includegraphics[width=0.12\linewidth]{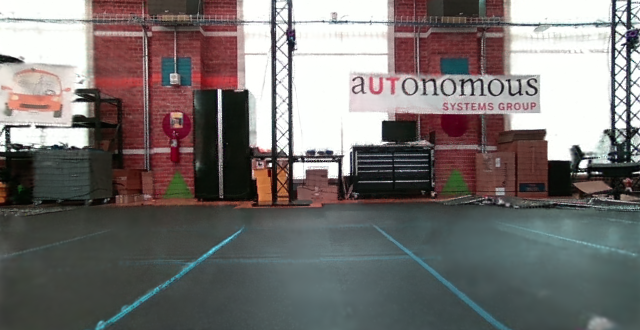} & \includegraphics[width=0.12\linewidth]{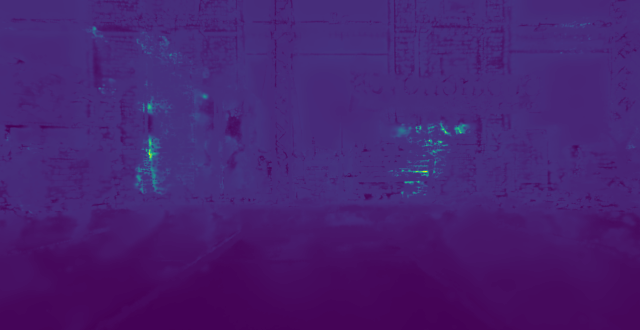} \\
    & \multicolumn{2}{c}{Square-1} & \multicolumn{2}{c}{Ego-centric-1} & \multicolumn{2}{c}{Ego-drive} & \multicolumn{2}{c}{Fast-straight}
    \end{tabular}
    \caption{Qualitative results on UT-MM dataset:~~ 
    RGB and depth renderings of UT-MM scenes. Note that the ground truth (GT) depths are captured with depth cameras, and thus are imperfect. Our method exhibits geometric details not present in the GT depth, as well as fewer RGB artifacts compared to SplaTAM. }
    \label{fig:renderresults}
\end{figure*}

\begin{table*}[ht]
  \centering
  \caption{Multi-modal SLAM results on the UT-MM dataset:~~ATE RMSE $\downarrow$ is in cm and PSNR $\uparrow$ is in dB, with SplaTAM is used as a baseline. Best results are in \textbf{bold}. Both depth and inertial measurements benefit tracking and image quality. }
  \begin{tabular}[c]{ccccccccccc}
    \toprule
    Method
    & \multicolumn{2}{c}{Avg}
    & \multicolumn{2}{c}{Square-1}  
    & \multicolumn{2}{c}{Ego-centric-1} 
    & \multicolumn{2}{c}{Ego-drive} 
    & \multicolumn{2}{c}{Fast-straight} \\
    
    \cmidrule(r){2-3} \cmidrule(r){4-5} \cmidrule(r){6-7} \cmidrule(r){8-9} \cmidrule(r){10-11}
    & ATE & PSNR & ATE & PSNR & ATE & PSNR & ATE & PSNR & ATE & PSNR \\ \midrule

    SplaTAM (RGB-D) & 12.06 & 22.03 & 32.86 & \textbf{18.67} & 4.40 & 22.78 & \textbf{4.20} & 20.61 & 6.78 & 26.07 \\ 
    Ours (RGB) & 39.14 & 19.73 & 59.48 & 16.54 & 4.09 & 23.151 & 67.20 & 17.51 & 25.78 & 21.71 \\ 
    Ours (RGB+IMU) & 33.23 & 19.58 & 44.26  & 17.01 & 3.41& 22.96 & 68.50 & 17.12 & 16.78 & 21.24 \\ 
    Ours (RGB-D) & 8.75 & 22.20 & 20.38 & 16.55 & 6.86 & 22.24 & 4.25 & 23.58 & 3.52 & \textbf{26.42} \\ 
    \textbf{Ours} (RGB-D+IMU) & \textbf{3.98} & \textbf{23.30} & \textbf{7.11} & 18.59 & \textbf{1.15} & \textbf{24.95} & 4.54 & \textbf{23.61} & \textbf{3.13} & 26.05 \\ 
    \bottomrule
  \end{tabular}
  \label{tab:UT-MMresults}  
\end{table*}

In a real-time setting, it is impractical to optimize the 3D Gaussian map over the entire set of video frames. However, one can exploit the temporal locality of video frames to prune away most frames from the optimization pool. This process of selectively choosing a subset of frames to optimize is known as keyframing. 

In general, keyframes should be chosen to minimize the number of redundant frames processed and maximize the information gain. MM3DGS achieves this by selecting an input frame as a keyframe when the map does not contain enough information to track the current frame. This is calculated using a covisibility metric, which defines which Gaussians are visible in multiple keyframes.

To do so, first, a depth rendering is created using the current frame's estimated pose. This depth rendering can be backprojected into a point cloud. This point cloud can be projected onto the image plane of any keyframe's estimated pose. The covisibility can then be defined as the percentage of points visible in the keyframe. If this covisibility drops below 95\%, the current frame is added as a keyframe.

In the presence of visual noise, such as when an input frame depicts motion blur, keyframes with low image quality may persist and degrade the reconstruction and tracking quality. To prevent outlier noisy images from being selected, the Naturalness Image Quality Evaluator (NIQE) metric \cite{mittal2013niqe} is used to select the highest quality frame across a sliding window.

During the mapping phase, Gaussians are optimized over the set of keyframes covisible with the current frame. In this way, the optimization affects all of the relevant measurements available while minimizing processing of redundant frames and reducing computational load.

\subsection{Mapping}

The mapping process optimizes the Gaussian features visible within the current covisible set of keyframes. For the current frame and each selected keyframe, the 3D Gaussians are optimized according to

\begin{equation}\label{eq:mapping}
    \mathcal{L_\mathrm{mapping}} = \lambda_\mathrm{C}\mathcal{L_\mathrm{photo}} + \lambda_\mathrm{S}\mathcal{L_\mathrm{SSIM}} + \lambda_\mathrm{D} \mathcal{L_\mathrm{depth}}
\end{equation}

\noindent where $\mathcal{L}_\mathrm{SSIM}$ is the D-SSIM loss \cite{wang2004ssim}. The other terms are identical to those in Eq. (\ref{eq:tracking}). To prevent unconstrained growth of Gaussians into unobserved areas, Gaussians are constrained to be isotropic.

\section{EXPERIMENTAL SETUP}

\subsection{Datasets}

Several visual-intertial SLAM datasets exist to test visual-inertial fusion, however they provide grayscale images rather than RGB images, which fails to test reconstruction capabilities \cite{Burri25012016eurocmav, Schubert_2018tumvi}. An exception is the AR Table dataset, which contains RGB images along with inertial measurements. However, it is limited to egocentric scenes focused at only tables \cite{chen2023artable}. To bridge this gap due to the absence of RGB visual-inertial datasets, we release such a dataset, dubbed UT Multi-modal (UT-MM), captured in the Anna Hiss Gymnasium at the University of Texas at Austin. A Clearpath Jackal unmanned ground vehicle (UGV), shown in \cref{fig:jackal}, is equipped with a Lord Microstrain 3DM-GX5-25 IMU, an Intel Realsense D455 camera, and an Oster 64 line LiDAR. The UGV captures RGB and depth images at 30 frames per second, inertial measurements at 100 Hz and point clouds at 10 Hz. In addition, the ground truth trajectory of the UGV is captured using a 16 camera Vicon motion capture system with sub-mm precision. A view from one of the 16 cameras is shown in \cref{fig:jackal}.

\begin{figure}[t]
    \centering
    \includegraphics[width=0.8\linewidth]{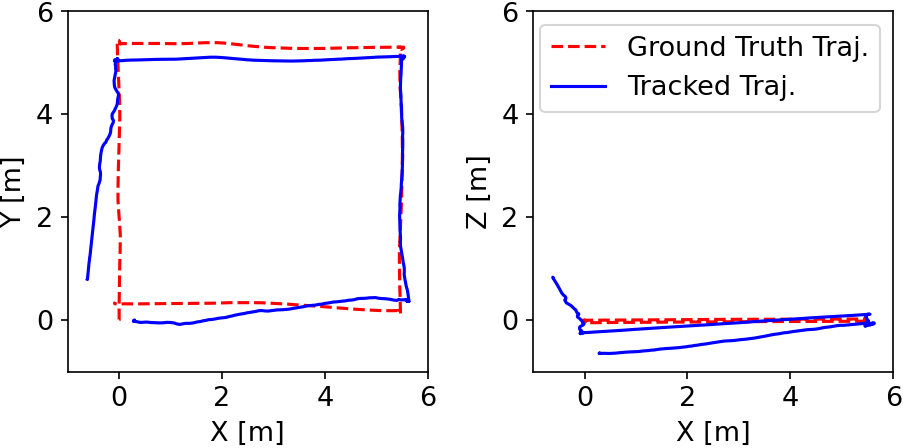}
    \includegraphics[width=0.8\linewidth]{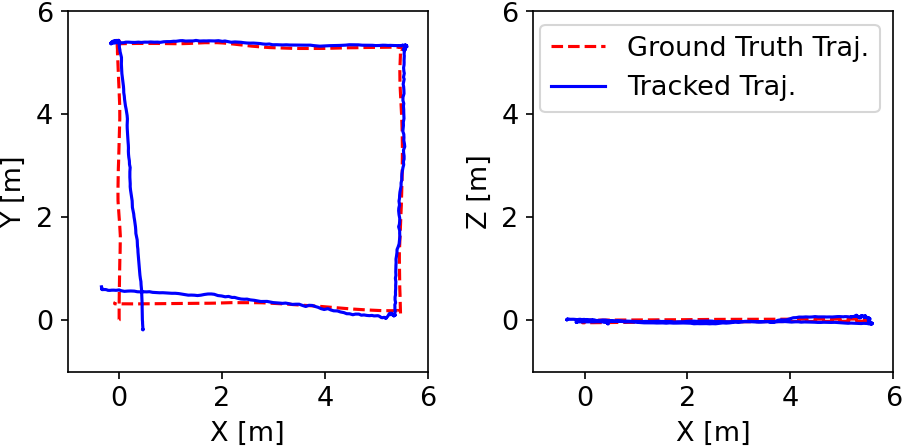}
    \includegraphics[width=0.8\linewidth]{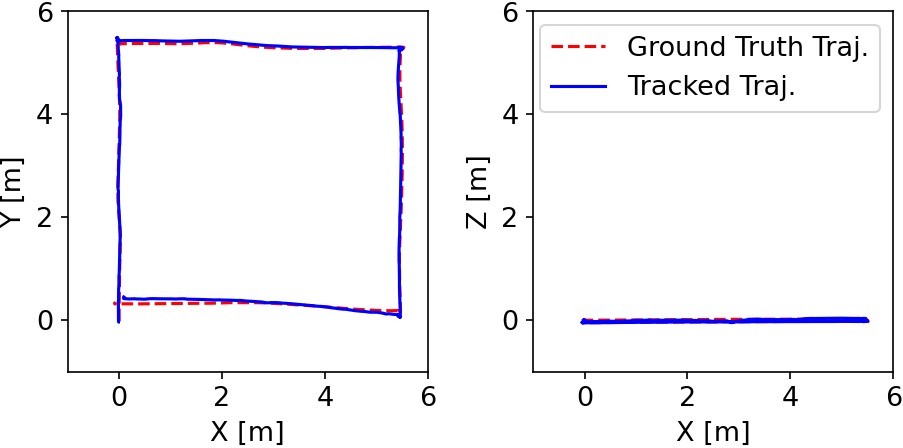}
    \caption{Tracking results for the UT-MM Square-1 scene. The blue solid line denotes the tracked trajectory, while the red dotted line denotes the ground truth. Top: monocular RGB case exhibits substantial drift. Middle: RGB-D case fixes Z drift, but XY drift persists. Bottom: Adding IMU measurements to RGB-D fixes XY drift.}
    \label{fig:multimodalsquare}
\end{figure}

\begin{figure}
    \centering
    \includegraphics[width=0.8\linewidth]{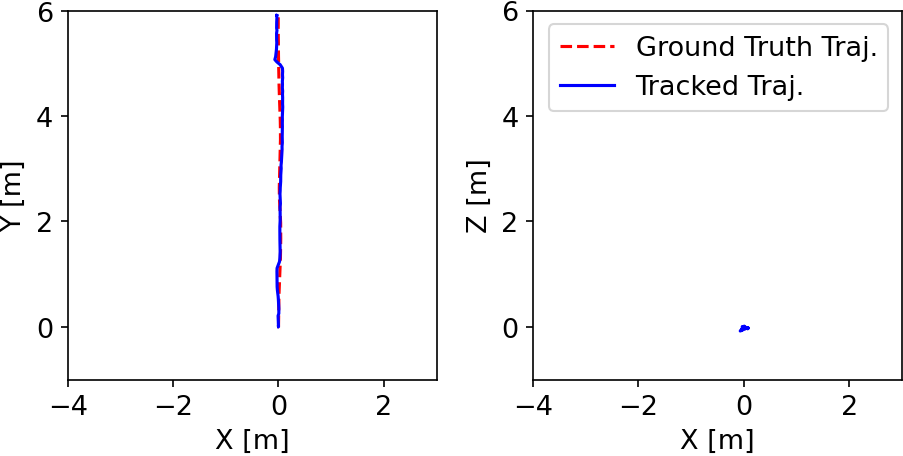}
    \includegraphics[width=0.8\linewidth]{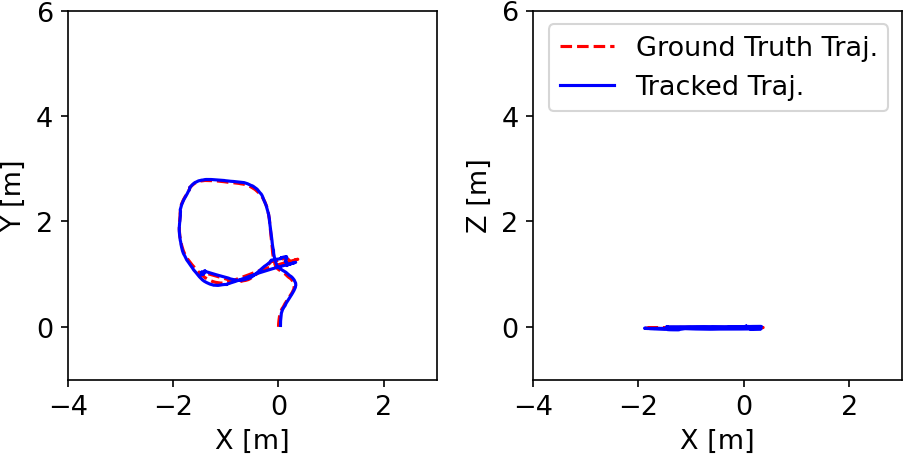}
    \includegraphics[width=0.8\linewidth]{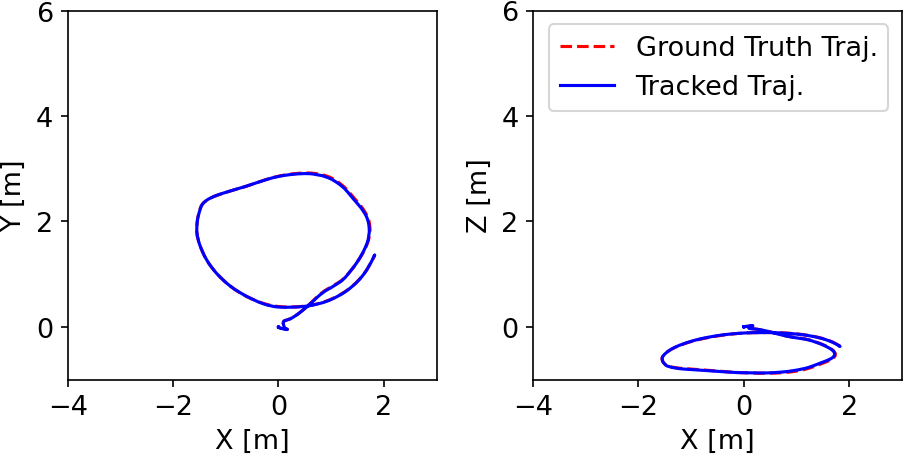}
    \caption{RGB-D+IMU tracking trajectory results for selected scenes in the UT-MM dataset. Top: Fast-straight. Middle: Ego-drive. Bottom: Ego-centric-1.}
    \label{fig:multimodalscenes}
\end{figure}

The UT-MM dataset contains eight different scenes labeled: Ego-centric-1, Ego-centric-2, Ego-drive, Square-1, Square-2, Fast-straight, Slow-straight-1, and Slow-straight-2. The three straight scenes are short scenes with the Jackal driving along a straight path with different initial positions and speeds. Square-1 and Square-2 consist of straight driving with three turns to create a loop depicting a square-shaped trajectory. Ego-drive consists of the robot being driven around several obstacles, capturing the obstacles from 360$\degree$. The Ego-centric-1 and Ego-centric-2 datasets were captured with the Jackal mounted on a omnidirectional trolley revolving around an object of interest while keeping it focused nearly at the center of the image. In some scenes, such as in the beginning of Ego-centric-1, dynamic objects are also present. These may either be cropped out or be included to test the robustness of a SLAM method to dynamic objects.

As shown in \cref{fig:UT-MM}, there are several sensing modalities available in the UT-MM dataset. For each frame in the dataset, a corresponding RGB image and depth map is available. At the same time, IMU measurements are provided at a rate of 100 Hz along with accurate point clouds captured. RGB-D images and IMU measurements are software-synchronized; an additional hardware synchronization board takes Pulse Per Second (PPS) input to synchronize LiDAR and IMU measurements. Our framework does not make use of highly accurate LiDAR measurements given the high cost of the sensor and lack of easy access by a common user. Nonetheless, the LiDAR can be used as ground truth during evaluation of a depth estimator.

In addition to UT-MM, we test our framework on the TUM RGB-D dataset to evaluate the performance of the monocular SLAM model in a different setting \cite{sturm12tumrgbd}. Note that TUM RGB-D provides accelerometer data but no gyroscope measurements and thus is not suitable for 6-DOF inertial fusion.

\subsection{Metrics}
To evaluate our model, we use two main metrics: 1) tracking accuracy, measured via absolute tracking error root mean square error (ATE RMSE), and 2) scene reconstruction quality, measured via peak signal-to-noise ratio (PSNR). An Umeyama point alignment algorithm aligns the trajectory generated using the model of interest with the ground truth trajectory \cite{umeyama1991least}. SplaTAM, an RGB-D 3DGS SLAM model that this work extends, is used as a baseline \cite{keetha2023splatam}.

\subsection{Implementation}
We run our framework on an RTX A5000 GPU with 24GB VRAM. We do not multi-thread our framework and hence the tracking and mapping threads run sequentially. For all scenes, we run pose optimization for 100 iterations followed by 150 iterations of map optimization. We set the loss function hyperparameters to $\lambda_{\mathrm{C}}=0.8$, $\lambda_{\mathrm{S}}=0.2$, and $\lambda_{\mathrm{D}}=0.05$.

\section{RESULTS AND DISCUSSION}

We conduct a comprehensive qualitative and quantitative evaluation of our framework on the UT-MM dataset and perform an ablation study on the TUM RGB-D dataset to justify the use of NIQE keyframing and Pearson correlation loss.

\subsection{Effect of Incorporating Multi-modal Sensor Information}

\begin{figure}
    \centering
    \includegraphics[width=0.8\linewidth]{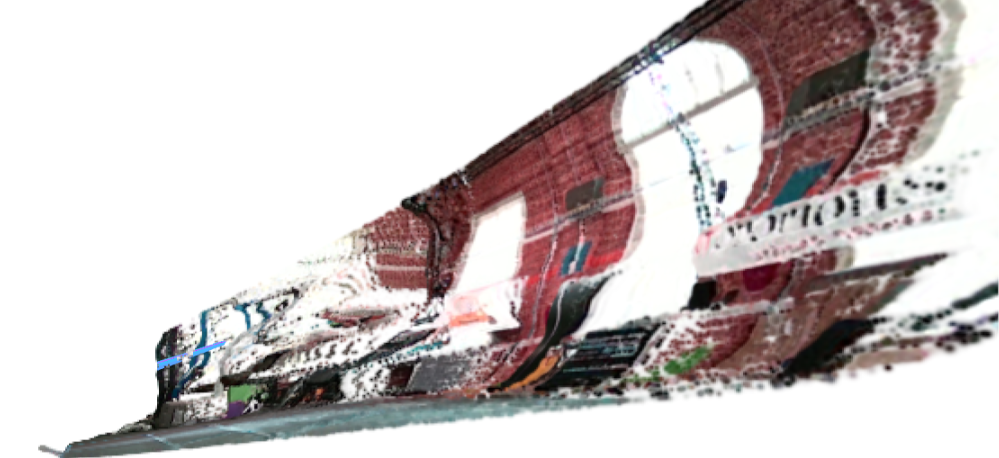}
    \includegraphics[width=0.8\linewidth]{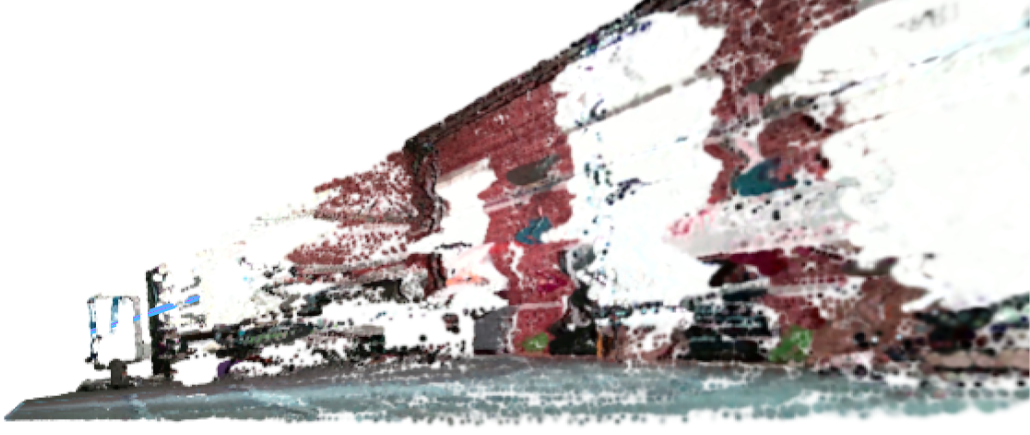}
    \caption{Top: depth initialized by DPT. Bottom: depth initialized by depth camera. The DPT estimate exhibits warping along the Z axis compared to the depth camera.}
    \label{fig:dptwarp}
\end{figure}

\Cref{tab:UT-MMresults} shows the tracking and rendering results for SplaTAM, a state-of-the-art RGB-D 3DGS baseline, and various configurations of MM3DGS on scenes from the UT-MM dataset. Considering the average ATE and PSNR metrics, it is evident that addition of inertial measurements improves tracking and image quality for both the RGB and RGB-D configurations. The RGB-D+IMU configuration consistently outperforms others, demonstrating a \textbf{3x} improvement in ATE RMSE and \textbf{5\%} improvement in PSNR compared to the baseline on average. This demonstrates the value of integrating inertial measurements in our framework enhancing trajectory tracking and rendering quality as previously claimed.

Furthermore, we showcase qualitative comparisons of the rendered RGB and depth images against SplaTAM and ground truth in Fig. \ref{fig:renderresults}. The superior image rendering quality of our framework on all of the scenes compared to SplaTAM signifies the value of the RGB-D+IMU configuration of MM3DGS. Due to the use of lightly weighted depth correlation loss rather than a heavily weighted direct depth loss implemented in SplaTAM, MM3DGS is able to capture geometric details that are not present in the noisy depth input through the photometric loss in \cref{eq:photoloss}. For instance, in the Ego-centric-1 scene, our depth rendering includes the many small holes on the back of the chair while SplaTAM instead overfits to the depth input and is not able to capture the holes. MM3DGS also exhibits significantly fewer visual artifacts in its RGB renderings compared to SplaTAM which depicts considerably more missing colors and floaters. Both methods render at 90 fps on an RTX A5000 GPU.

To visualize the effect of incorporating IMU and depth measurements on tracking, we plot the trajectory of the RGB, RGB-D, and RGB-D+IMU configurations on the UT-MM Square-1 scene as shown in \cref{fig:multimodalsquare}. 
The integration of depth measurements helps with trajectory alignment by eliminating drift in the Z-axis. The dense depth estimates provided by DPT are indeed warped in the Z direction as shown in \cref{fig:dptwarp}. However, drift along the XY plane persists and is addressed only with the addition of inertial measurements in our framework. This highlights the limitations of monocular depth estimators and the importance of integrating both inertial and depth sensing modalities.

\begin{figure}[t]
    \setlength\tabcolsep{0.5pt}
    \adjustboxset{width=\linewidth,valign=c}
    \centering
    \begin{tabular}{ccccccccc}
    \rotatebox[origin=B]{90}{w/o Pearson corr.}  & \includegraphics[width=0.49\linewidth]{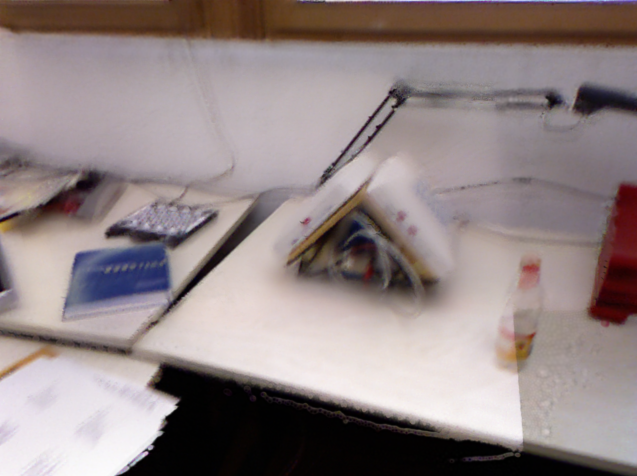} & \includegraphics[width=0.49\linewidth]{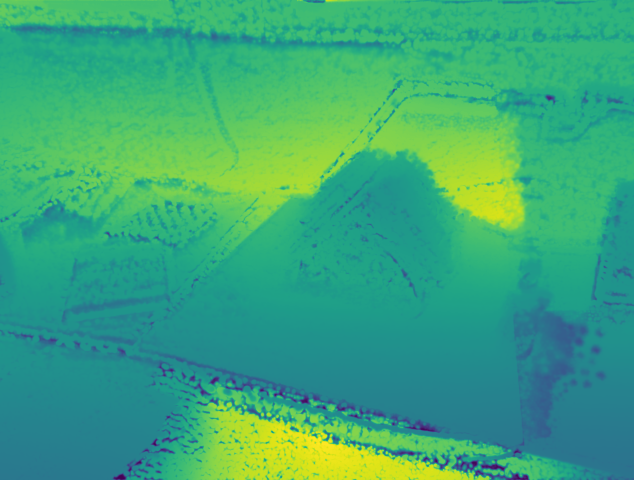} \\
    \rotatebox[origin=B]{90}{w/ Pearson corr.}  & \includegraphics[width=0.49\linewidth]{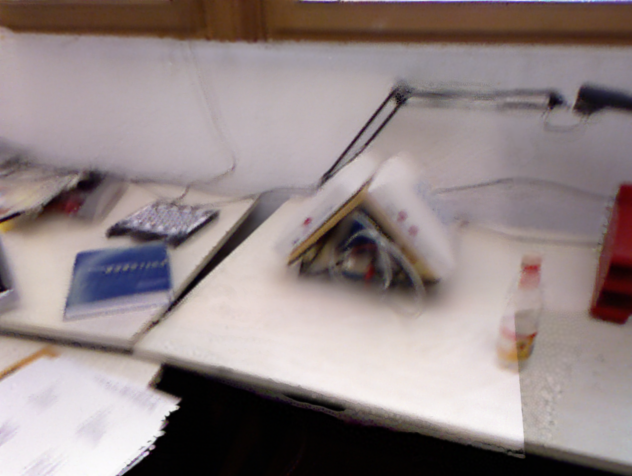} & \includegraphics[width=0.49\linewidth]{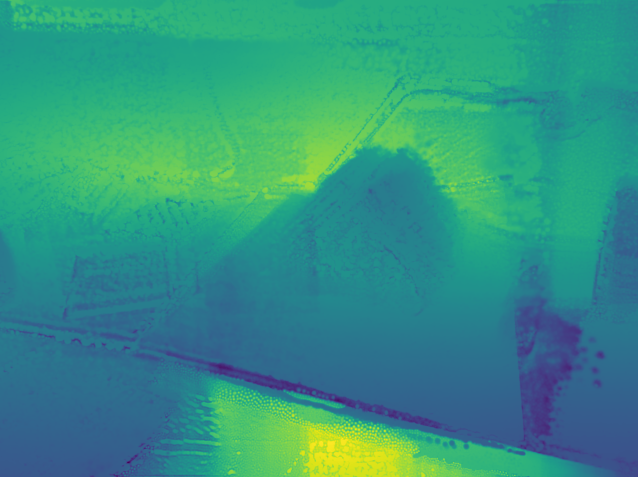} \\
    \end{tabular}
    \caption{RGB and depth renderings of TUM RGB-D with and without Pearson correlation loss. The addition of Pearson correlation results in similar RGB quality while increasing geometric consistency, particularly along edges. Note the shift in brightness on the right side of the RGB renders; this is due to a sudden exposure change in the input camera frames.}
    \label{fig:pearsonablation}
\end{figure}

\begin{table}[!ht]
  \centering
  \caption{Monocular RGB configuration results on the TUM RGB-D dataset. ATE RMSE $\downarrow$ is in cm. Monocular RGB SLAM provides comparable performance as RGB-D baselines.}
  \begin{tabular}[c]{cccc}
    \toprule
    Method
    & fr1/desk  
    & fr1/desk2 
    & fr2/xyz \\
    \midrule

    SplaTAM (RGB-D) & \textbf{3.35} & 6.54 & \textbf{1.24} \\
    \textbf{Ours} (RGB) & 3.51 & \textbf{5.78} & 2.04 \\
    \bottomrule
  \end{tabular}
  \label{tab:tumresults}  
\end{table}

\begin{table}[!ht]
  \centering
  \caption{Monocular RGB configuration ablation results on the TUM RGB-D {\tt freiburg1/desk2} scene. ATE RMSE $\downarrow$ is in cm and PSNR $\uparrow$ is in dB. Performance is best with both Pearson Corr. loss and NIQE keyframing.}
  \begin{tabular}[c]{cccc}
    \toprule
    NIQE Keyframing
    & Pearson Corr. Loss
    & ATE RMSE
    & PSNR \\
    \midrule

    \cmark & \xmark & Fails & Fails \\
    \xmark & \cmark & 7.8 & 18.05 \\
    \cmark & \cmark & \textbf{5.78} & \textbf{18.33} \\
    \bottomrule
  \end{tabular}
  \label{tab:ablation}  
\end{table}

\subsection{Performance of the Monocular RGB Configuration}

To further gauge the performance of the proposed monocular RGB SLAM configuration, we evaluate its performance on the TUM RGB-D dataset. The ATE RMSE results on select scenes from the dataset are shown in \cref{tab:tumresults}. In ego-centric scenes, the RGB model exhibits comparable performance to SplaTAM's RGB-D model, demonstrating that depth measurements do not provide much added information given a wide range of multi-view constraints. However, the RGB model fails on scenes that involve longer trajectories which may be alleviated by the addition of loop closure methods.

\subsection{Ablation Study}

We also perform an ablation study on the {\tt freiburg1/desk2} scene due to presence of high motion blur and exposure changes. \Cref{tab:ablation} showcases RGB tracking results with Pearson correlation depth loss and NIQE keyframing ablated.
Without the Pearson correlation depth loss, tracking fails. The effect of the depth correlation loss in improving geometric consistency is illustrated in \cref{fig:pearsonablation}. Further, the addition of NIQE keyframing increases both tracking accuracy and image quality.

\section{CONCLUSIONS}

We presented MM3DGS, a multi-modal SLAM framework built on a 3D Gaussian map representation that utilizes visual, inertial, and depth measurements to enable real-time photorealistic rendering and improved trajectory tracking. We evaluate our framework on a new multi-modal dataset, UT-MM, that includes RGB-D images, IMU measurements, LiDAR depth, and ground truth trajectories. MM3DGS achieves superior tracking accuracy and rendering quality compared to state-of-the-art baselines. In addition, we present an ablation study to highlight the importance of our framework. MM3DGS can be implemented in a wide range of applications in robotics, augmented reality, and mobile computing due to its use of commonly available and inexpensive sensors. As future work, MM3DGS can be extended to include tightly-coupled IMU fusion and loop closure to further enhance tracking performance. 

\addtolength{\textheight}{-12cm}   









\bibliographystyle{IEEEtran}
\bibliography{refs}

\end{document}